\documentclass[letterpaper]{article} 
\usepackage[arxiv]{aaai24}
\usepackage{times}  
\usepackage{helvet}  
\usepackage{courier}  
\usepackage[hyphens]{url}  
\usepackage{graphicx} 
\usepackage{natbib}  
\usepackage{caption} 
\usepackage{algorithm}
\usepackage{algorithmic}

\usepackage{newfloat}
\usepackage{listings}
\DeclareCaptionStyle{ruled}{labelfont=normalfont,labelsep=colon,strut=off} 
\floatstyle{ruled}
\newfloat{listing}{tb}{lst}{}
\floatname{listing}{Listing}
\usepackage{subcaption}
\usepackage{booktabs}

\usepackage{amsmath}
\usepackage{amsthm}
\usepackage{amssymb}
\usepackage{mathtools}
\usepackage{bm}
\usepackage{mathrsfs}
\usepackage{newtxmath}

\theoremstyle{plain}
\newtheorem{theorem}{Theorem}

\newtheorem{example}{Example}
\theoremstyle{definition}
\newtheorem{definition}{Definition}
\newtheorem{assumption}{Assumption}
\theoremstyle{remark}

\def\cF{{\mathcal{F}}}
\def\cG{{\mathcal{G}}}

\def\cM{{\mathcal{M}}}

\def\bbM{{\mathbb{M}}}

\newcommand{\op}[1]{\operatorname{#1}}

\title{DIGIC: Domain Generalizable Imitation Learning by Causal Discovery}

\author {
    Yang Chen\textsuperscript{\rm 1},
    Yitao Liang\textsuperscript{\rm 2},
    Zhouchen Lin\textsuperscript{\rm 1}
}
\affiliations {
    \textsuperscript{\rm 1}School of Intelligence Science and Technology, Peking University\\
    \textsuperscript{\rm 2}Institute for Artificial Intelligence, Peking University\\
    \{yangchen, yitaol, zlin\}@pku.edu.cn
}

\begin{document}

\maketitle

\begin{abstract}
Causality has been combined with machine learning to produce robust representations for domain generalization.
Most existing methods of this type require
massive data from multiple domains to identify causal features by cross-domain variations, which can be expensive or even infeasible and may lead to misidentification in some cases.
In this work, we make a different attempt by leveraging the demonstration data distribution to discover the causal features for a domain generalizable policy. We design a novel framework, called DIGIC, to identify the causal features by finding the direct cause of the expert action from the demonstration data distribution via causal discovery.
Our framework can achieve domain generalizable imitation learning with only single-domain data and serve as a complement for cross-domain variation-based methods under non-structural assumptions on the underlying causal models. Our empirical study in various control tasks shows that the proposed framework evidently improves the domain generalization performance and has comparable performance to the expert in the original domain simultaneously.
\end{abstract}

\section{Introduction}

Imitation learning is a widespread paradigm for policy making problems in various real-world applications, such as autonomous vehicles, robotics, healthcare, and dialogue systems. By learning from expert demonstrations, imitation learning aims to train a policy to emulate the expert policy that generates demonstrations. Many real-world applications require that learned policies are generalizable for different domains \citep{zhang2020invariant, zhang2020learning}. For example, a driving policy for autonomous vehicles is expected to work in all cities and areas. A fundamental challenge in this field is enabling the policy to generalize and perform comparably to the expert across unseen or new domains, a concept known as domain generalization in imitation learning.

\begin{figure}[t]
    \begin{subfigure}{\linewidth}
        \centering
        \includegraphics[width=\linewidth]{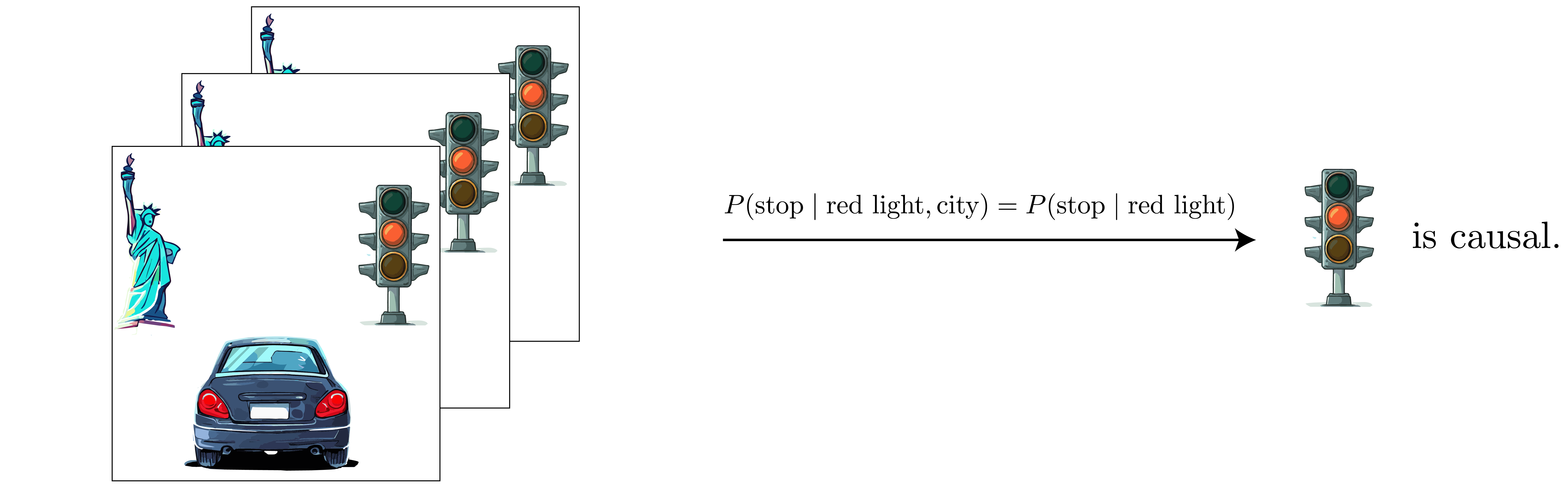}
        \caption{Distribution-based methods.}
        \label{fig:sd}
    \end{subfigure}
    \begin{subfigure}{\linewidth}
        \centering
        \includegraphics[width=\linewidth]{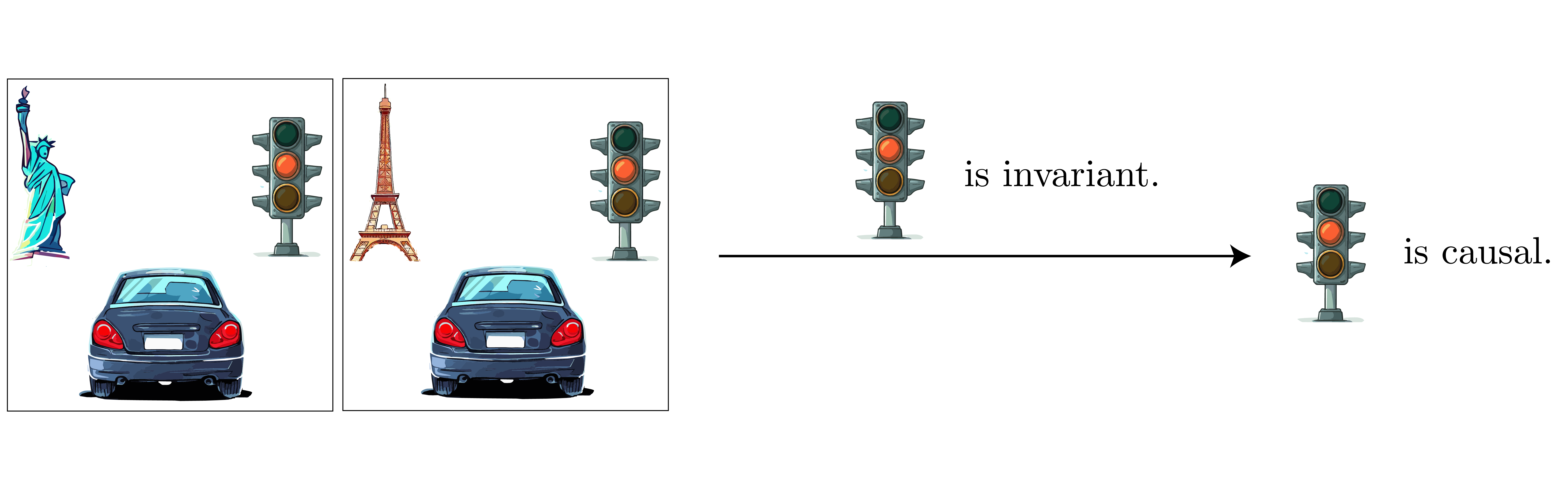}
        \caption{Cross-domain variation-based methods.}
        \label{fig:md}
    \end{subfigure}
    \caption{Various principles to identify causal features. In (a), distribution-based methods leverage the conditional independence relations (i.e., the stopping action is independent of the city conditioned on the traffic light) in the demonstration data distribution to identify that the traffic light is the direct cause of the stopping action. In (b), cross-domain variation-based methods identify the traffic light as the causal feature because it is invariant across domains.}
    \label{fig:sd-md}
\end{figure}

To achieve domain generalization, we need to understand the causal mechanism behind expert decisions \citep{scholkopf2021toward}. This mechanism remains consistent even with domain shifts. Therefore, we should build an imitation policy that follows this same causal mechanism. More specifically, this can be achieved by identifying the causal features of the expert action and conditioning the imitation policy on these features.
Existing methods for causal feature identification often rely on multi-domain data to leverage variations across domains, recognizing the invariant components as causal features. However, these multi-domain methods identify causal features based on invariance rather than truly understanding the causal structure. As a result, they are non-causal methods in nature, despite their wide applications in causal feature identification. This approach encounters significant challenges when multi-domain data is scarce or even unavailable. Moreover, when the variation across multi-domain data fails to fully capture the true causal components, these methods might misidentify the essential features, leading to suboptimal domain generalization. To illustrate this point, consider the following example:
\begin{example}\label{ex:toy}
    We are to find causal features among three covariates $X_1$, $X_2$, and $X_3$ for $A$.
    Suppose that we have training data from two domains whose distributions are $P_1(A, X_1, X_2, X_3)$ and $P_2(A, X_1, X_2, X_3)$, respectively. We further suppose $P_1(X_2=x_2)=1$, $P_2(X_2=x'_2)=1$ for some $x_2\neq x'_2$, and 
    \begin{equation*}
        \begin{aligned}
            &P_1(A, X_1, X_3 \mid X_2=x_2) 
            =P_2(A, X_1, X_3 \mid X_2=x'_2)\\
            &=P(A, X_1, X_3)
            =P(A\mid X_3)P(X_1\mid X_3)P(X_3).        
        \end{aligned}
    \end{equation*}
    In this case, $X_1$ and $X_2$ are the spurious features, and $X_3$ is the causal feature.
    The variation across the domains can only identify $X_2$ as a spurious feature but falsely recognize $X_1$ as a causal feature 
    because $P(A,X_1,X_3)$ is invariant across domains.
\end{example}

In this work, we propose a novel approach to address the domain generalization problem of imitation learning by integrating a causal discovery module within the imitation learning framework. 
Our method exploits the causal structure between variables directly from the data distribution (Figure~\ref{fig:sd-md} provides an intuitive illustration of the difference between distribution-based methods and cross-domain variation-based methods).  
We introduce a \textbf{D}oma\textbf{I}n \textbf{G}eneralizable \textbf{I}mitation Learning by \textbf{C}ausal Discovery (DIGIC) framework that combines causal discovery with imitation learning. 
For each imitation learning task, we construct a causal model using the demonstration distribution. 
We then identify the variables corresponding to the direct cause of the expert action as the causal features. 
Such causal features lead to a domain generalizable imitation policy across various domains, 
because the policy has a consistent causal relationship between the covariates and the action with the expert 
and thus replicate the expert behavior across any domain where the causal structure holds.

The essence of our method lies in extracting the structure between the variables from the demonstration data distributions. 
Specifically, we identify a subset of covariates that directly cause the expert action by analyzing the conditional independence relations within the data. 
These covariates are unique in that they remain correlated with the action even when conditioned on any other subset, while all other covariates become independent of the action when conditioned on this identified subset. 
By employing causal discovery techniques, we distill this critical causal structure from the demonstration data distribution. 
Then, by training an imitation policy to match the distribution of the action conditioned on these direct causes, we aim to replicate the expert policy precisely, 
leveraging the inherent causality to achieve a more robust and generalizable solution. For example, consider an autonomous vehicle trained to navigate in one city by identifying universal traffic principles such as stopping at red lights, on which the human driving policy depends. By using causal discovery to discern these causal rules from the data distribution, it extracts relationships that are conditionally independent of city-specific features. This enables the vehicle to generalize its driving policy to unseen domains, i.e., other cities, and achieve comparable performance as the human driving policy, effectively demonstrating how domain generalization is feasible by utilizing the conditional independence relations in the data distribution.
Our method, focusing on the causal structure derived from data distribution, also serves as a powerful remedy for cross-domain variation-based methods.
Utilizing the causal structure in the data distribution itself, our method can find spurious features that are invariant across domains.
In Example~\ref{ex:toy}, the variation across domains fails to remove the spurious feature $X_1$ while the independence between $A$ and $X_1$ conditioned on $X_3$ reflects that $X_1$ is a spurious feature. 

Our main contributions are summarized as follows:
\begin{itemize}
    \item We introduce the DIGIC framework, a novel approach that integrates causal discovery to identify causal features from demonstration data distribution for domain generalization. 
     This method can achieve domain generalizable imitation policy with single-domain data and compensate for the shortcomings of cross-domain variation-based methods when the cross-domain variation is not sufficient to reflect all spurious features.
    \item We develop a methodology in imitation learning that exploits the properties of the demonstration data distribution to learn a domain generalizable policy. This methodology goes beyond the prevailing paradigm that utilizes cross-domain variations. Our method uncovers insights to foster generalizable policies without the need for cross-domain variations, widening its applicability.
    \item We introduce an innovative fusion of causality and machine learning in the realm of imitation learning. 
    Our research elucidates methods for leveraging causal discovery techniques to augment machine learning algorithms with flexibility.
\end{itemize}

\section{Related Work}
\textbf{Causal Imitation Learning.}
Imitation learning aims to mimic expert behavior given a set of demonstrations. 
Behavior Cloning (BC) \citep{pomerleau1988alvinn} and its variants \citep{ross2010efficient,ross2011reduction,duan2017one} learn an imitation policy by directly approximating the map from observations to actions of the expert with supervised learning. 
Another line of algorithms roots from Inverse Reinforcement Learning (IRL) \citep{russell1998learning, abbeel2004apprenticeship, ziebart2008maximum}, 
which learns a reward function from the demonstration and utilizes the learned reward function for planning. 

Causal inference has been combined with imitation learning in recent years to refine analysis \citep{zhang2020causal, kumor2021sequential} or address some bottlenecks \citep{de2019causal, bica2021invariant, lu2022invariant} in imitation learning. 
Our work tackles the domain generalization problem in imitation learning by explicitly discovering the underlying causal model of expert data generation.

\textbf{Causal Discovery.}
The goal of causal discovery is to reconstruct the underlying SCM given observational or interventional data generated by the causal model \citep{spirtes2000causation, glymour2019review, vowels2022d}.
Typical general-purpose causal discovery algorithms include constraint-based methods \citep{spirtes1995directed, zhang2011kernel, spirtes2000causation} and score-based methods \citep{geiger1994learning, chickering2002optimal, huang2018generalized}.
Apart from these two types, some algorithms can solve causal discovery problems more efficiently 
by making stronger assumptions \citep{shimizu2006linear, hoyer2008nonlinear, xie2020generalized} or finding weaker models than SCMs \citep{lauritzen1996graphical, loh2013structure, chandrasekaran2012latent, taeb2018interpreting}. 
The learning-based causal discovery algorithm used in this work is inspired by the graph estimation algorithm based on generalized inverse covariance matrix \citep{loh2013structure}. 

\textbf{Distributional Robustness.}
The domain generalization problem considered in this paper can be categorized as a problem of distributional robustness, the ability of a model against distributional shift \citep{ben2009robust, duchi2021statistics}. To achieve distributional robustness, some optimize the worst-case loss among a class of distributions \citep{delage2010distributionally, duchi2021learning},
some exploit data from multiple domains to find invariant representations \citep{arjovsky2019invariant, ahuja2020invariant},  and others also encode inductive biases according to domain knowledge \citep{higgins2016beta, cobbe2019quantifying, zhang2020learning}.

In recent years, there have been attempts to utilize causal models of problems for distributional robustness \citep{peters2016causal, scholkopf2021toward, sun2021recovering}, most of which assume the structures of the causal models to some extent.
Our work also exploits causal models for distributional robustness.
Unlike previous approaches, our method uniquely leverages data distribution to pinpoint causal features, eliminating the need for cross-domain variations and operating under minimal non-structural assumptions.

One of the most related works to ours is \citet{de2019causal}, which also studies the out-of-distribution problem via a causality lens and does not require cross-domain variations in the training stage. However, our work is different from theirs in at least three aspects. First, our point is why and how one can achieve domain generalization and identify spurious features without cross-domain variations, while their point is why one should care about spurious features at all. Second, our work introduces a general framework (i.e., DIGIC) to combine causality and imitation learning for domain generalization, while their work proposes a specific method that trains a single graph-parameterized policy. Additionally, our work only needs observational data, while their work requires extra interventional data to select a final causal graph.

\section{Preliminaries}
We introduce the essential notations and concepts used in the paper in this section.

\textbf{Structural Causal Model.} A Structural Causal Model (SCM) $\cM$ is a tuple 
$\left\langle\bm U,\bm V, P(\bm U), \cF\right\rangle$ \citep{pearl2009causality}. $\bm U$ is the exogenous variables and $\bm V$ is the endogenous variables. $P(\bm U)$ represents the distribution of the exogenous variables $\bm U$. $\cF$ is the set of functionals that define the causal relations between variables. Each endogenous variable $V\in\bm V$ is generated by $f_{V}(\bm V',\bm U')$ for some $f_{V}\in\cF$, $\bm V'\subset\bm V$, and $\bm U'\subset\bm U$. The induced causal diagram $\cG$ of the SCM $\cM$ is a directed acyclic graph (DAG) whose nodes are the exogenous variables $\bm U$ and the endogenous variables $\bm V$. The directed edge $V_1 \rightarrow V_2$ is linked in $\cG$ if and only if there exists a functional $f_{V_2}\in\cF$ such that $V_2=f_{V_2}(\bm V', \bm U')$ and $V_1\in\bm V'\cup\bm U'$. We call $V_1$ a direct cause of $V_2$ if $V_1 \rightarrow V_2$ is an edge in $\cG$. Intervention can be described by the $do$-operator in the language of SCM. The experimental distribution of random variable $Y$ obtained by the intervention $\pi$ is represented as $P(Y\mid do(\pi))$. For further details on the SCM, refer to \citet{pearl2009causality}.

\textbf{Imitation Learning.}
Imitation learning aims to learn a policy from expert demonstrations. In this paper, we focus on the strict batch setting, i.e., all the demonstrations are collected in advance, and there is no interactive query to the expert during the training process. In strict batch imitation learning, we suppose that we have a set of expert demonstrations $\{(\bm o_i, \bm a_i)\}_{i=1}^N$. Each data point includes a recorded observation $\bm o$ and the corresponding expert action $\bm a$. Suppose the outcome of interest is $\bm y$. The goal is to learn a policy $\pi$ with the supervision of the expert demonstrations such that the distribution of the outcome $\bm y$ generated by the policy $\pi$ is the same as the one generated by the expert.

\begin{figure*}[t]
    \centering
    \includegraphics[width=\linewidth]{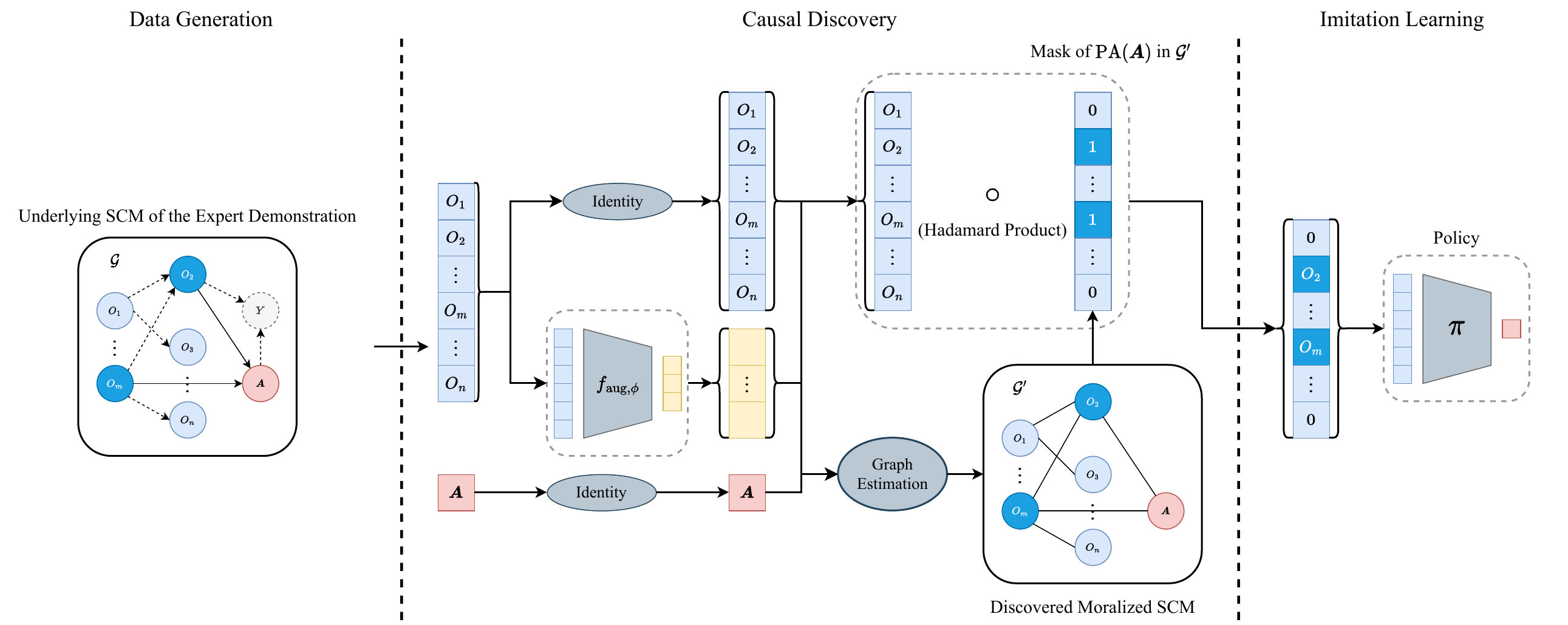}
    \caption{
    Overview of our DIGIC framework implementation. The training stage is comprised of two key modules: Causal Discovery and Imitation Learning. In Causal Discovery, the moralized structure is estimated using a generalized inverse based on the observations and the action, and the mask of $\op{PA}(\bm A)$ is extracted. The Hadamard product of $\bm O$ and this mask reveals causal features, which are employed by the policy net in the Imitation Learning module to derive a domain generalizable policy. During the inference stage, the causal discovery is bypassed, and the previously discovered mask of $\op{PA}(\bm A)$ is applied to the observations $\bm O$ with the resulting causal features fed into the policy model for decision-making.}
    \label{fig:framework}
\end{figure*}

\section{Method}
Our goal is to learn an imitation learning policy that can match the expert behavior across multiple domains. We formalize this domain generalization problem of imitation learning problem in terms of causal inference and propose DIGIC, a framework that incorporates causal discovery to solve this problem by extracting causal features from demonstration data distributions.

\subsection{Domain Generalization of Imitation Learning}\label{subsec:sdicil}
Suppose the expert demonstration is collected in an environment whose underlying SCM $\cM_0=\left\langle\bm U,\bm V, P_0(\bm U), \cF_0\right\rangle$. Denote the expert policy by $\pi_E=\pi_E(\bm A\mid \bm S)$, which is conditioned on a subset of the covariates $\bm S \subset \bm V$. Suppose the outcome of interest is $Y\in\bm V$. The distribution of $Y$ generated by the expert policy in $\cM_0$ is $P_{\cM_0}\left(Y\mid do(\pi_E)\right)$.
Suppose the collected demonstrations are $\{(\bm o_i, \bm a_i)\}_{i=1}^N$ and we can approximately regard the distribution $P_{\cM_0}\left(\bm O,\bm A\mid do(\pi_E)\right)$ as available in the analysis if the sample size $N$ is sufficiently large. 

We want an imitation policy that generalizes across multiple domains, i.e., the imitation policy always leads to the same outcome distribution as the expert in all possible domains. Definition~\ref{def:dgip} states the concept of domain generalizable imitation policy in this paper formally.
\begin{definition}[Domain Generalizable Imitation Policy]\label{def:dgip}
 Let $Y$ be the outcome of interest and $\bbM$ be the set of SCMs representing all potential environments. We say an imitation policy $\pi=\pi(\bm A\mid \bm 
 X)$ is domain generalizable with respect to the expert policy $\pi_E$ and the set of SCMs $\bbM$ if for all $\cM\in\bbM$
 \begin{equation}\label{eq:iip}
     P_{\cM}\left(Y\mid do(\pi)\right) = P_{\cM}\left(Y\mid do(\pi_E)\right),
 \end{equation}
where $\bm X$ is a subset of the endogenous variables.
\end{definition}

While most existing methods require multi-domain data and leverage cross-domain variations to distinguish the causal and non-causal features, our work proposes to exploit the demonstration data distribution to extract the causal features directly by causal discovery techniques to attain a domain generalizable policy, which can apply with only single-domain data and do not rely on cross-domain variations.

\subsection{The DIGIC Framework}

We present the DIGIC framework, which integrates causal discovery with imitation learning to derive a domain generalizable policy. The core principle of DIGIC is to identify the direct cause of the action in the SCM of the demonstration. This approach leverages the conditional independence relations of the demonstration data, using the direct cause of the expert action in the SCM as the basis for imitation. Rather than simply mimicking the expert's behavior in specific scenarios, the DIGIC policy captures the underlying causal mechanisms that govern the expert's decision-making. Provided sufficient data, it fits the action distribution conditioned on the direct cause and approximates the exact expert policy, leading to a domain generalizable policy that has comparable performance to the expert in different domains. Importantly, as DIGIC does not depend on cross-domain variations, it is suitable for single-domain applications and offers a valuable remedy when multi-domain methods are not applicable.

The DIGIC framework unfolds in two stages: causal discovery and imitation learning. In the causal discovery stage, a causal model is constructed from the training data, and the direct cause of the expert action is identified as the causal representation for imitation. Following this, the imitation learning stage crafts an imitation policy conditioned on this causal representation. Further details of these stages are elaborated below. Moreover, we detail an experimental implementation of DIGIC that employs a learning-based causal discovery module, utilizing a generalized inverse covariance matrix. This two-stage process underscores the unique contribution of DIGIC, offering a robust and concise framework for domain generalizable imitation learning.

\subsubsection{Causal Discovery}
We begin by constructing a causal model using the training samples, recognizing that the recovery of a precise SCM with only observational data is generally unattainable. In fact, the SCM can be recovered only up to a Markov equivalent class, even with a precise distribution of observational data \citep{peters2017elements}. The breakthrough lies in the realization that it is sufficient to discover the direct cause efficiently by estimating a graphical model. This model is weaker than the original SCM but retains the essential information required for domain generalizable imitation learning.

Given the nature of decision-making in imitation learning, where covariates are observed prior to any action, it is unnecessary to delineate the complete SCM. Instead, we aim to identify the covariates directly correlated with the action. To accomplish this, we recover the undirected moral graph of the SCM, which can be inferred from observational data \citep{wainwright2008graphical}. This moral graph, obtained by connecting the parents of each child and undirecting all edges, accurately reflects the direct correlation between the direct cause and the action (see the example in the appendix for further illustration). The neighbors of the expert action $\bm A$ within this moralized SCM represent the direct cause in the original SCM, and thus recovering the undirected moral graph suffices for our objectives.

\subsubsection{Imitation Learning}
Upon discovering the direct cause through the causal discovery process, it is then used as a robust feature in the imitation learning stage. This approach is not limited to a specific algorithm; although we use Behavioral Cloning (BC) for simplicity, the framework can integrate with various offline imitation algorithms.

This two-step approach ensures a precise correlation between the action and its direct cause, facilitating the creation of a robust imitation policy that is applicable across various domains from the demonstration data distribution. By focusing on direct causality rather than an intricate SCM, DIGIC maintains efficiency without compromising its foundational goal of domain generalization in imitation learning.

\subsubsection{Implementation}
We implement the DIGIC framework in a differentiable pipeline, using a learning-based approach for causal discovery. Specifically, we adopt a variant of the generalized inverse covariance algorithm, making use of neural networks to enable efficient application beyond original restrictions \citep{lauritzen1996graphical, loh2013structure, zhang2018ista, xie2019differentiable}.

For the causal discovery module, we design a learning-based variant of the graph estimation method based generalized inverse covariance matrix. Previous work has shown that under some conditions, the inverse covariance matrix $\Sigma^{-1}$ or its generalized variant, i.e., the inverse covariance matrix of certain augmented variables, is graph-structured, i.e., $\Sigma^{-1}_{ij}\neq 0$ if and only if the undirected edge ${i,j}$ is linked in the moralized SCM \citep{lauritzen1996graphical, loh2013structure}. Our method attempts to extend the properties to general distributions. We train a neural network to construct augmented variables such that their inverse covariance matrix is graph-structured. More specifically, we train a neural network $f_{\text{aug},\theta}(\bm o)$ such that the inverse covariance matrix of the random vector 
\begin{equation}
    \bm v = \bm o \odot \bm a\odot\bm f_{\text{aug},\theta}\left(\bm o \odot \bm a\right).
\end{equation}
reflects the graph structure of the moralized SCM.
We then estimate the generalized inverse covariance matrix of the augmented vector. Let $\{(\bm o_i, \bm a_i)\}_{i=1}^{N}$ be a batch of data and $\{\bm v_i\}_{i=1}^N$ be the augmented vectors. The empirical estimation of the generalized covariance matrix $\Hat{\Sigma}$ is
\begin{equation}\label{eq:emp-cov}
    \Hat{\Sigma} = \sum_{i=1}^N v_i \left(v_i\right)^\intercal - 
    \left(\sum_{i=1}^N v_i\right)\left(\sum_{i=1}^N v_i\right)^\intercal.
\end{equation}
We estimate the generalized inverse covariance matrix by the inverse matrix $\Hat{\Sigma}^{-1}$.

For imitation learning, we employ a simple BC policy network, utilizing the discovered causal features as representation. The whole pipeline of the DIGIC framework is differentiable with respect to learnable parameters. During the training stage, the entire model is trained end-to-end to minimize the imitation loss. During the inference stage, we fix a precomputed causal graph based on a training data batch. This combined with the causal graph is used to construct the representation for the final imitation policy model. 

Figure~\ref{fig:framework} illustrates our implementation of the DIGIC framework. The strength of DIGIC also lies in its flexibility; besides the above implementation, it can seamlessly integrate with various causal discovery algorithms and imitation learning methodologies. This adaptability enables customization for real-world applications, allowing practitioners to select the most appropriate causal discovery and imitation learning techniques to enhance domain generalization performance. We emphasize the general DIGIC framework in this work and leave exploring the impact of different causal discovery techniques and imitation learning methods on real-world problems for future research.

\section{Analysis}
In this section, we analyze our framework and show that the learned policy does generalize across domains under several non-structural assumptions. All the proofs are delayed to the appendix. We provide empirical validation for the domain generalization ability of our framework when the theoretical assumptions may not hold strictly in the next section 

\begin{assumption}[Interventional Domain Shift]\label{as:ids}
Each potential domain corresponds to an intervention on the observable variables of the original domain, i.e., for each $\cM\in\bbM$ and the SCM $\cM_0$ of the original domain,
there exists an intervention $\omega$ disjoint of $\bm A$ and $Y$ on the $\bm O^{\cM_0}$ such that $P_\cM\left(\cdot\right)=P_{\cM_0}\left(\cdot\mid do(\omega)\right)$.
\end{assumption}

Assumption~\ref{as:ids} describes domain shifts in terms of causal intervention.
This generalizes the common assumption that attributes 
domain shift to the distributional shift of the domain-specific noise variables.

\begin{assumption}[Coverage of Direct Cause]\label{as:coverage}
    For each $\cM\in\bbM$ and the SCM $\cM_0$ of the original domain, the support set of the margin distribution of the direct cause in the original domain contains that in the shifted domain, i.e.,
    $\op{supp}\left(P_{\cM}\left(\bm Z_d\right)\right)\subseteq\op{supp}\left(P_{\cM_0}\left(\bm Z_d\right)\right)$, where $\bm Z_d$ is the direct cause of the expert action.
\end{assumption}

Assumption~\ref{as:coverage} requires a partial coverage condition on the marginal distribution of the direct cause. This is much weaker than common assumptions that assume coverage of all the covariates \citep{spencer2021feedback, chang2021mitigating}.

\begin{assumption}[Faithfulness \citep{spirtes2000causation}]\label{as:faithfulness}
Let $\cG$ be a causal graph and $P$ be the distribution generated by $\cG$.
We say that $\langle G, P\rangle$ satisfies the faithfulness condition if and only if 
every conditional independence relation in $P$ is entailed by the causal graph $\cG$.
\end{assumption}

\begin{figure*}[t]
    \begin{subfigure}{0.33\linewidth}
        \centering
        \includegraphics[width=0.99\linewidth]{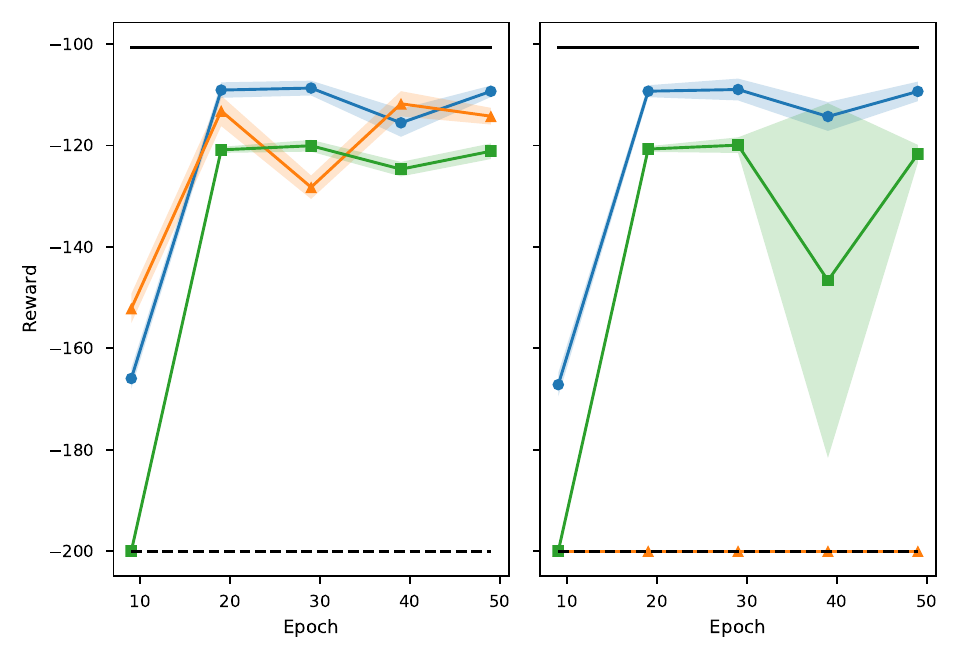}
        \caption{MountainCar-v0}
    \end{subfigure}
    \begin{subfigure}{0.33\linewidth}
        \centering
        \includegraphics[width=0.99\linewidth]{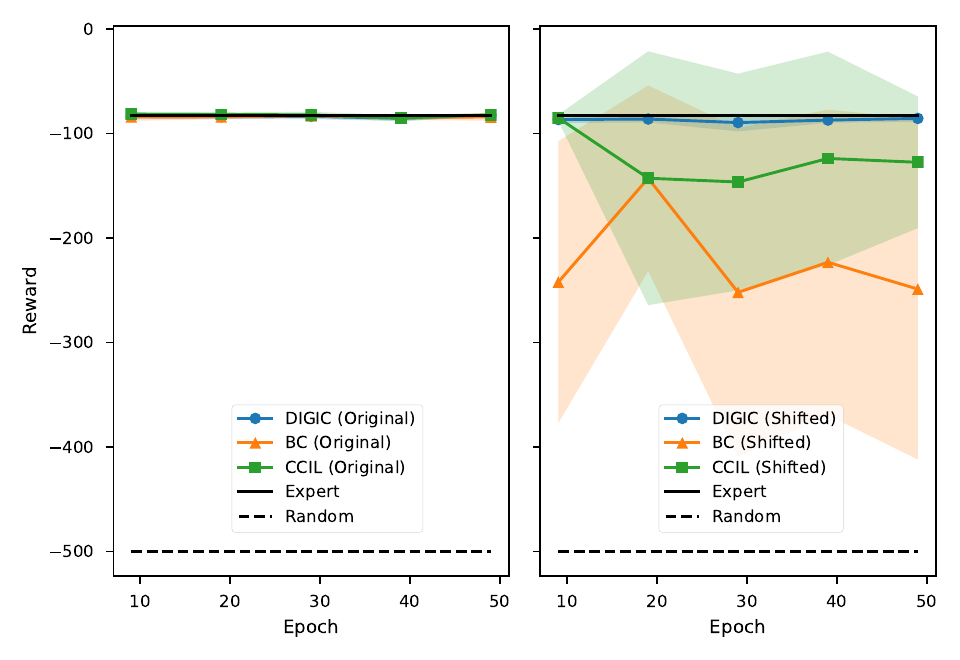}
        \caption{Acrobot-v1}
    \end{subfigure}
    \begin{subfigure}{0.33\linewidth}
        \centering
        \includegraphics[width=0.99\linewidth]{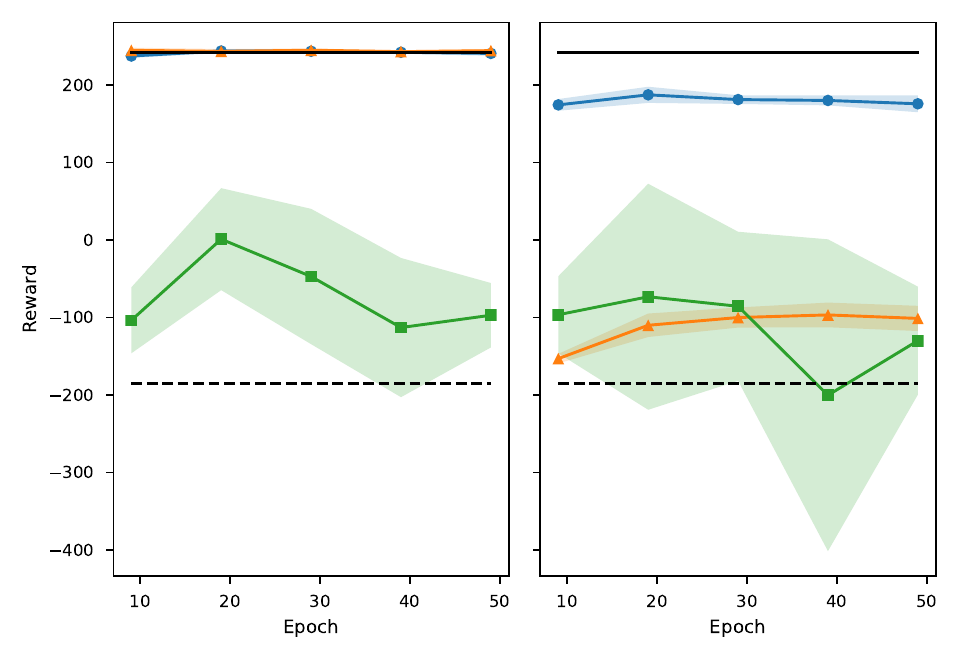}
        \caption{LunarLander-v1}
    \end{subfigure}
    \begin{subfigure}{0.33\linewidth}
        \centering
        \includegraphics[width=0.99\linewidth]{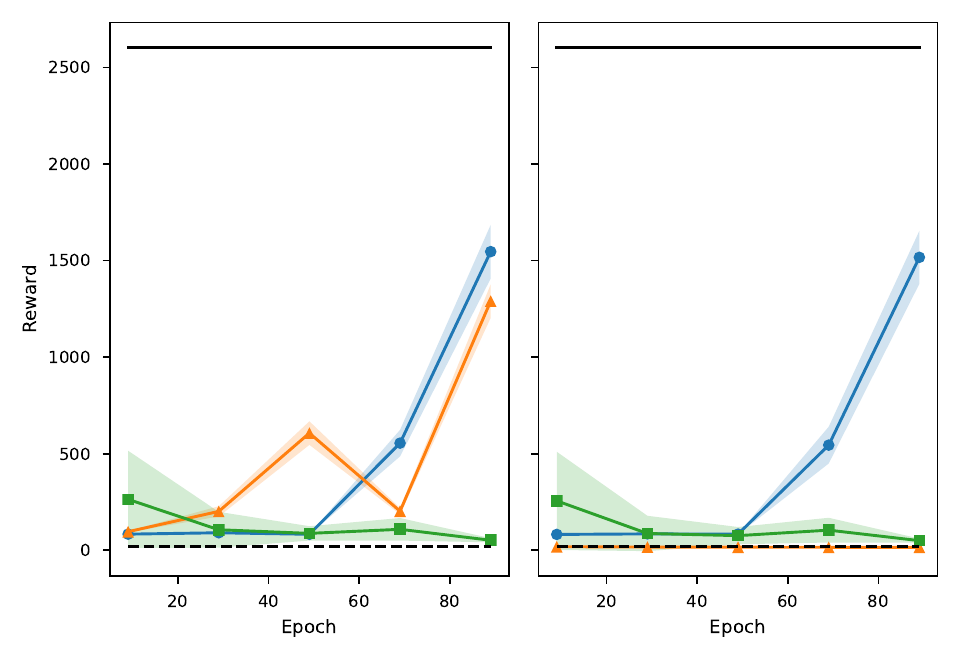}
        \caption{HopperBulletEnv-v0}
    \end{subfigure}
    \begin{subfigure}{0.33\linewidth}
        \centering
        \includegraphics[width=0.99\linewidth]{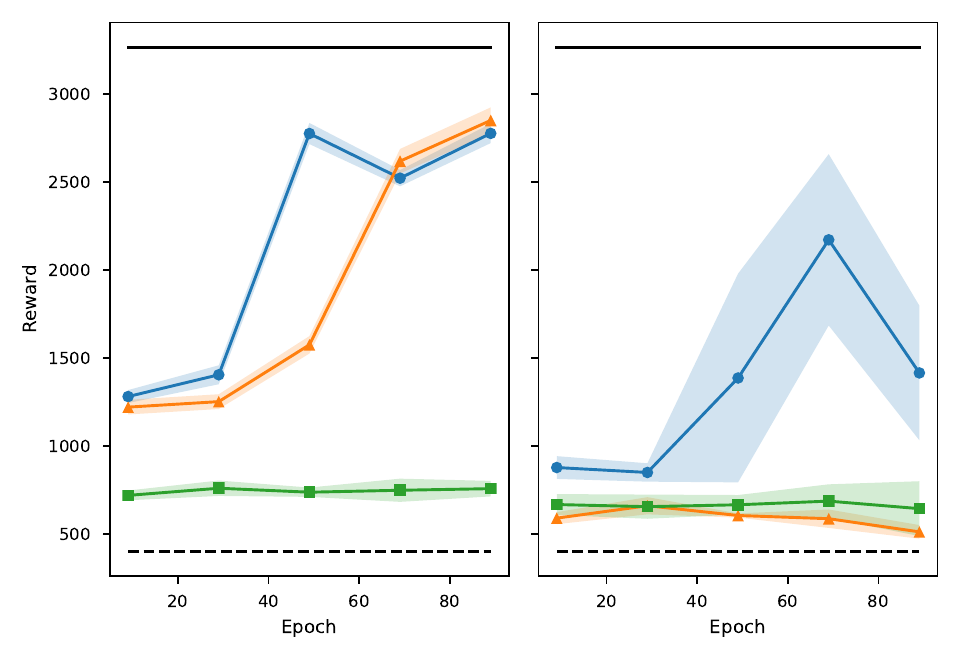}
        \caption{AntBulletEnv-v0}
    \end{subfigure}
    \begin{subfigure}{0.33\linewidth}
        \centering
        \includegraphics[width=0.99\linewidth]{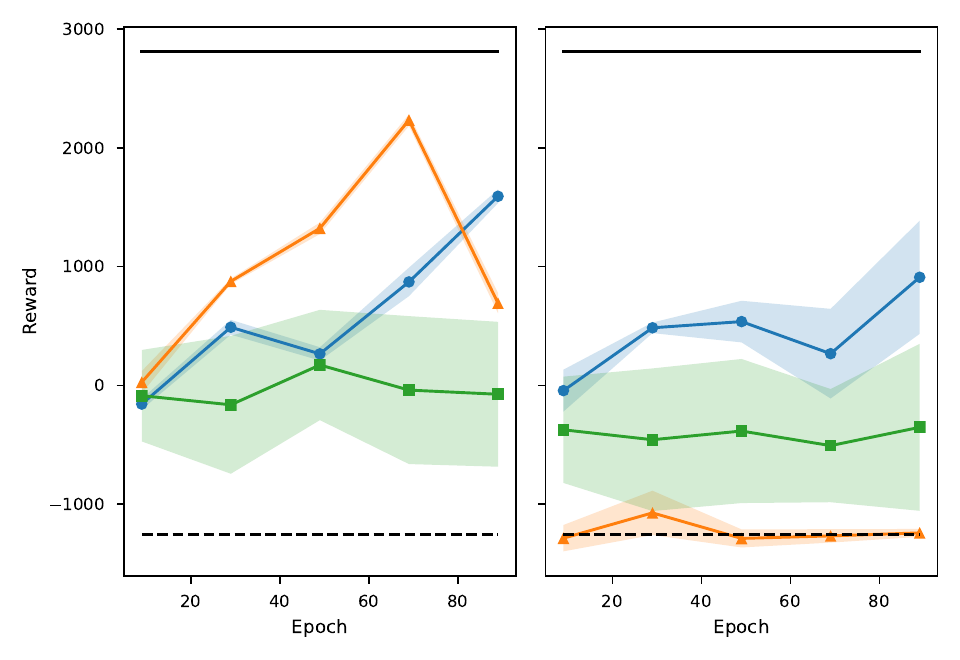}
        \caption{HalfCheetahBulletEnv-v0}
    \end{subfigure}
    \caption{Evaluation results of single-domain generalization in different tasks in the original and shifted domains. Each curve shows the mean and the standard deviation of the average total rewards over 10 trials. In each subfigure, the left part is the evaluation results in the original domain, and the right part is the evaluation results in the shifted domain. The $x$-axis indicates the training epoch at which the checkpoint is saved. The $y$-axis indicates the average total reward. We evaluate the checkpoints at different epochs. The legend is shared by all the subfigures. We see that DIGIC achieves comparable performance with the expert in most tasks when evaluated in the original domains and outperform BC and CCIL evidently in the shifted domains.}
    \label{fig:sd-results}
\end{figure*}
\begin{figure*}[t]
    \centering
    \begin{subfigure}{0.33\linewidth}
        \centering
        \includegraphics[width=0.99\linewidth]{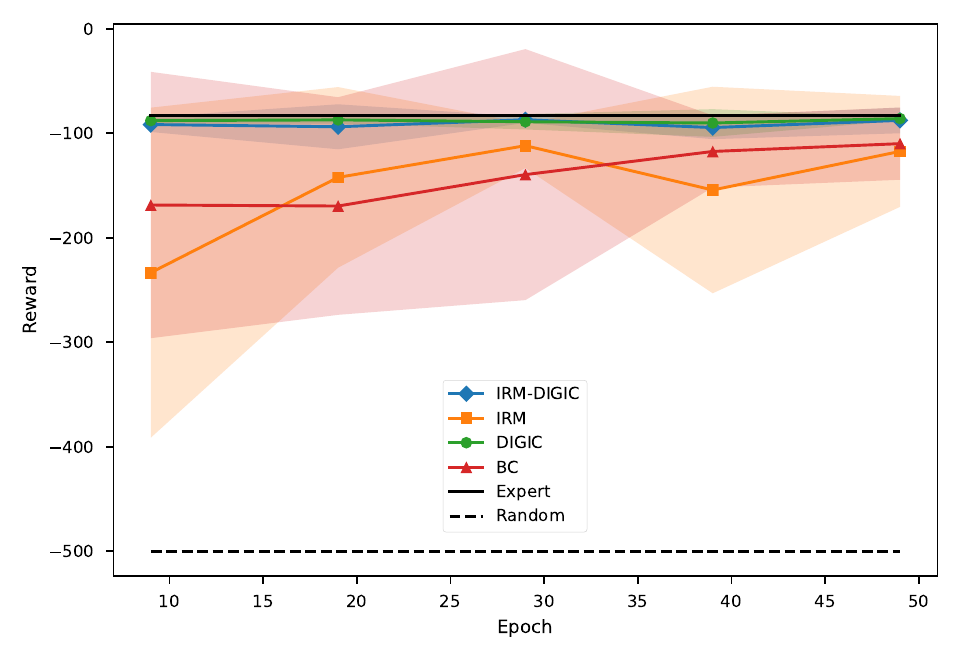}
        \caption{Acrobot-v1}
    \end{subfigure}
    \begin{subfigure}{0.33\linewidth}
        \centering
        \includegraphics[width=0.99\linewidth]{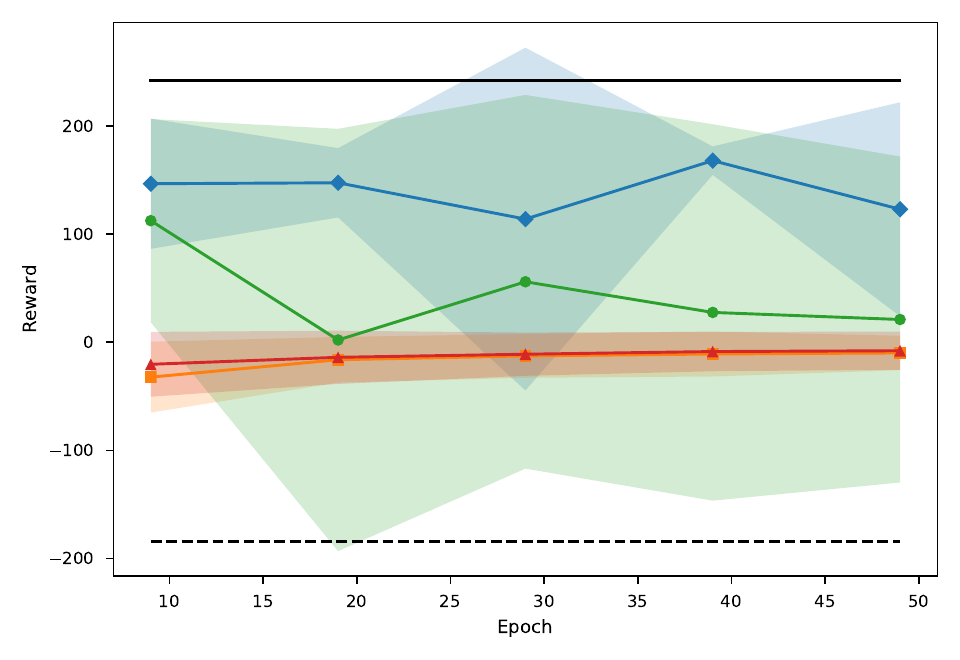}
        \caption{LunarLander-v1}
    \end{subfigure}
    \caption{Comparison of domain generalization performance with the invariant spurious features in two tasks. Each curve represents the mean and standard deviation of average total rewards across 10 trials, with the $x$-axis showing the training epoch and the $y$-axis indicating the average total reward. Checkpoints are evaluated at different epochs. The legend is shared by all the subfigures. The result reveals that IRM-DIGIC outperforms IRM evidently in the presence of the invariant spurious features.}
    \label{fig:md-results}
\end{figure*}

Assumption~\ref{as:faithfulness} is a common assumption in the literature of causal discovery, which allows building causal models from observational distributions. It is the key assumption under which it is possible to identify the causal mechanism of the expert decisions solely from the demonstration data distribution.
It is ubiquitous in real-world scenarios and the set of unfaithful models has Lebesgue measure zero \citep{spirtes2000causation, uhler2013geometry}.


We now concentrate on analyzing the causal imitation learning policy, assuming an ideal causal discovery model that identifies the precise moralized SCM. In our framework, we utilize the direct cause of the expert action, linked to the expert action's neighbors in the moralized SCM, as the covariates for imitation. Theorem~\ref{thm:iip} shows that the behavior cloning policy conditioned on these causal features ensures domain generalization.

\begin{theorem}\label{thm:iip}
Suppose that $\bm X \subset \bm O$ is the direct cause of the expert action $\bm A$ in the original SCM $\cM_0$. If Assumption~\ref{as:ids} and Assumption~\ref{as:coverage} hold, then the behavior cloning policy (at the population level)
\begin{equation}\label{eq:bc}
    \pi\left(\bm A\mid\bm X\right) = P_{\cM_0}\left(\bm A\mid\bm X\right)
\end{equation}
is a domain generalizable policy with respect to $\bbM$.
\end{theorem}

\section{Experiments}
We evaluate the DIGIC framework under two scenarios: single-domain generalization and enhancement for multi-domain methods. In single-domain generalization, we train imitation policies using data from one domain and test them in a shifted domain. In the enhancement for multi-domain methods, we train imitation policies with data spanning multiple domains, where cross-domain variations, however, do not readily distinguish causal from non-causal features. Both experiments require the methods to identify causal features when the cross-domain variation principle does not apply. We are to show the empirical effectiveness of the proposed DIGIC for domain generalization without cross-domain variations in real-world applications where the theoretical assumptions may not hold strictly.

\subsection{Single-Domain Generalization}

In single-domain generalization, we are unable to rely on cross-domain variations. Policies must exploit the underlying data distribution to achieve generalization. We assess the DIGIC framework on six control tasks in OpenAI Gym, constructing observations from different domains as described by \citet{bica2021invariant}. Our setup includes two spurious features: a domain-specific multiplicative factor, and a domain identifier. Training data is generated from a single domain, and the imitation agent is tested in a domain with different augmentation factors.

We contrast DIGIC with BC and CCIL \citep{de2019causal}, using expert and random policies as baselines. We evaluate the methods in both original and shifted domains, with an emphasis on cross-domain generalization. Further details on implementation and hyperparameters are provided in the appendix.

Our evaluation of policies, executed in the environments over 100 trajectories, is repeated 10 times per task (Figure~\ref{fig:sd-results}). The findings show that DIGIC generally surpasses BC and CCIL in unseen domains concerning the average total reward. DIGIC's performance is mostly on par with experts in original domains, with CCIL exhibiting better domain generalization than BC but failing in specific tasks. Our results demonstrate a substantial improvement in single-domain generalization performance with DIGIC, confirming its ability to leverage the causal structure of the data distribution to transcend the cross-domain variation principle.

Intuitively, the multiplicative factor's correlation with true causal features makes its identification more challenging than that of the domain identifier. Our causal discovery module confirms this, successfully identifying domain identifiers in all tasks but only recognizing multiplicative factors in the Hopper and Half Cheetah tasks. Table~\ref{tab:mask} summarizes these results. While the results demonstrate our method's effectiveness in distinguishing most features, they also uncover a limitation in identifying features strongly correlated with true causal ones, which confirms the initial intuition.

\begin{table}[h]
    \begin{center}
    \begin{tabular}{lcc}
    \toprule
    Task            & Mult Factor          & Env Id \\
    \midrule
    Mountain Car    & $0 / 1$      & $1 / 1$ \\
    Acrobot         & $0 / 3$      & $1 / 1$ \\
    Lunar Lander    & $0 / 3$      & $1 / 1$ \\
    Hopper          & $1 / 1$      & $1 / 1$ \\
    Ant             & $1 / 3$      & $1 / 1$ \\
    Half Cheetah    & $4 / 5$      & $1 / 1$ \\
    \bottomrule
    \end{tabular}
    \end{center}
    \caption{Spurious feature identification results of DIGIC, where ``Mult Factor'' means ``Multiplicative Factor'' and ``Env Id'' means ``Environment Identifier''. The entry ``$a / b$'' in the table means $a$ dimensions out of the total $b$ dimensions of the spurious feature are identified. }
    \label{tab:mask}
\end{table}

\subsection{Enhancement for Multi-Domain Methods}

We test DIGIC's capability to enhance multi-domain methods on two control tasks. The construction of training data is similar to the single-domain experiment but includes an additional invariant spurious feature. Such a feature may lead to substantial domain generalization degradation for policies learned by multi-domain methods.

Our comparisons include IRM, a prototypical multi-domain method, DIGIC, IRM-DIGIC (an enhanced version of IRM with DIGIC), and BC, with expert and random policies as baselines. We assess the domain generalization performance, with details and settings provided in the appendix.

Figure~\ref{fig:md-results} summarizes the results. IRM only achieves domain generalization comparable to BC in the presence of the invariant spurious features, revealing that traditional multi-domain methods can suffer from significant performance degradation in domain generalization. IRM-DIGIC exhibits the best performance, illustrating DIGIC's clear enhancement of IRM for invariant spurious features. The results of the enhancement for multi-domain methods against invariant spurious features provide empirical evidence that DIGIC can identify causal features for domain generalizable imitation learning directly from the demonstration data distribution, eliminating the need for cross-domain variations.

\section{Conclusion}
In this paper, we investigate domain generalization methods in causal imitation learning based on the statistical properties of demonstration data distributions instead of cross-domain variations. We clarify the formal definition of domain generalizable imitation learning in the language of causality and show the feasibility of learning a robust policy that generalizes across multiple domains conditioned on the causal features. Introducing the DIGIC framework, we tackle this challenge to identify the causal features directly from the training data distribution through causal discovery. This approach facilitates domain generalizable imitation learning by harnessing the inherent structure of the data distribution, eliminating reliance on cross-domain variations. 

There are several directions for future work.
The DIGIC framework offers a versatile approach to exploit causal discovery for domain generalization within imitation learning. Expanding the scope of DIGIC to encompass learning paradigms beyond imitation learning holds significant potential.
DIGIC underscores the potential of domain generalization methods rooted in data distribution, moving past the cross-domain variation principle. Delving into diverse statistical properties and methodologies could further amplify the potency of domain generalization techniques.

\bibliography{aaai24}

\begin{thebibliography}{48}
\providecommand{\natexlab}[1]{#1}

\bibitem[{Abbeel and Ng(2004)}]{abbeel2004apprenticeship}
Abbeel, P.; and Ng, A.~Y. 2004.
\newblock Apprenticeship learning via inverse reinforcement learning.
\newblock In \emph{International Conference on Machine Learning}, 1.

\bibitem[{Ahuja et~al.(2020)Ahuja, Shanmugam, Varshney, and
  Dhurandhar}]{ahuja2020invariant}
Ahuja, K.; Shanmugam, K.; Varshney, K.; and Dhurandhar, A. 2020.
\newblock Invariant risk minimization games.
\newblock In \emph{International Conference on Machine Learning}, 145--155.
  PMLR.

\bibitem[{Arjovsky et~al.(2019)Arjovsky, Bottou, Gulrajani, and
  Lopez-Paz}]{arjovsky2019invariant}
Arjovsky, M.; Bottou, L.; Gulrajani, I.; and Lopez-Paz, D. 2019.
\newblock Invariant risk minimization.
\newblock \emph{arXiv preprint arXiv:1907.02893}.

\bibitem[{Ben-Tal, El~Ghaoui, and Nemirovski(2009)}]{ben2009robust}
Ben-Tal, A.; El~Ghaoui, L.; and Nemirovski, A. 2009.
\newblock \emph{Robust optimization}, volume~28.
\newblock Princeton University Press.

\bibitem[{Bica, Jarrett, and van~der Schaar(2021)}]{bica2021invariant}
Bica, I.; Jarrett, D.; and van~der Schaar, M. 2021.
\newblock Invariant causal imitation learning for generalizable policies.
\newblock \emph{Advances in Neural Information Processing Systems}, 34:
  3952--3964.

\bibitem[{Chandrasekaran, Parrilo, and
  Willsky(2012)}]{chandrasekaran2012latent}
Chandrasekaran, V.; Parrilo, P.~A.; and Willsky, A.~S. 2012.
\newblock Latent variable graphical model selection via convex optimization.
\newblock \emph{The Annals of Statistics}, 40(4): 1935--1967.

\bibitem[{Chang et~al.(2021)Chang, Uehara, Sreenivas, Kidambi, and
  Sun}]{chang2021mitigating}
Chang, J.; Uehara, M.; Sreenivas, D.; Kidambi, R.; and Sun, W. 2021.
\newblock Mitigating covariate shift in imitation learning via offline data
  with partial coverage.
\newblock \emph{Advances in Neural Information Processing Systems}, 34:
  965--979.

\bibitem[{Chickering(2002)}]{chickering2002optimal}
Chickering, D.~M. 2002.
\newblock Optimal structure identification with greedy search.
\newblock \emph{Journal of Machine Learning Research}, 3(Nov): 507--554.

\bibitem[{Cobbe et~al.(2019)Cobbe, Klimov, Hesse, Kim, and
  Schulman}]{cobbe2019quantifying}
Cobbe, K.; Klimov, O.; Hesse, C.; Kim, T.; and Schulman, J. 2019.
\newblock Quantifying generalization in reinforcement learning.
\newblock In \emph{International Conference on Machine Learning}, 1282--1289.
  PMLR.

\bibitem[{De~Haan, Jayaraman, and Levine(2019)}]{de2019causal}
De~Haan, P.; Jayaraman, D.; and Levine, S. 2019.
\newblock Causal confusion in imitation learning.
\newblock \emph{Advances in Neural Information Processing Systems}, 32.

\bibitem[{Delage and Ye(2010)}]{delage2010distributionally}
Delage, E.; and Ye, Y. 2010.
\newblock Distributionally robust optimization under moment uncertainty with
  application to data-driven problems.
\newblock \emph{Operations research}, 58(3): 595--612.

\bibitem[{Duan et~al.(2017)Duan, Andrychowicz, Stadie, Jonathan~Ho, Schneider,
  Sutskever, Abbeel, and Zaremba}]{duan2017one}
Duan, Y.; Andrychowicz, M.; Stadie, B.; Jonathan~Ho, O.; Schneider, J.;
  Sutskever, I.; Abbeel, P.; and Zaremba, W. 2017.
\newblock One-shot imitation learning.
\newblock \emph{Advances in Neural Information Processing Systems}, 30.

\bibitem[{Duchi, Glynn, and Namkoong(2021)}]{duchi2021statistics}
Duchi, J.~C.; Glynn, P.~W.; and Namkoong, H. 2021.
\newblock Statistics of robust optimization: A generalized empirical likelihood
  approach.
\newblock \emph{Mathematics of Operations Research}, 46(3): 946--969.

\bibitem[{Duchi and Namkoong(2021)}]{duchi2021learning}
Duchi, J.~C.; and Namkoong, H. 2021.
\newblock Learning models with uniform performance via distributionally robust
  optimization.
\newblock \emph{The Annals of Statistics}, 49(3): 1378--1406.

\bibitem[{Geiger and Heckerman(1994)}]{geiger1994learning}
Geiger, D.; and Heckerman, D. 1994.
\newblock Learning gaussian networks.
\newblock In \emph{Uncertainty Proceedings 1994}, 235--243. Elsevier.

\bibitem[{Glymour, Zhang, and Spirtes(2019)}]{glymour2019review}
Glymour, C.; Zhang, K.; and Spirtes, P. 2019.
\newblock Review of causal discovery methods based on graphical models.
\newblock \emph{Frontiers in Genetics}, 10: 524.

\bibitem[{Higgins et~al.(2017)Higgins, Matthey, Pal, Burgess, Glorot,
  Botvinick, Mohamed, and Lerchner}]{higgins2016beta}
Higgins, I.; Matthey, L.; Pal, A.; Burgess, C.; Glorot, X.; Botvinick, M.;
  Mohamed, S.; and Lerchner, A. 2017.
\newblock beta-vae: Learning basic visual concepts with a constrained
  variational framework.
\newblock In \emph{International Conference on Learning Representations}.

\bibitem[{Hoyer et~al.(2008)Hoyer, Janzing, Mooij, Peters, and
  Sch{\"o}lkopf}]{hoyer2008nonlinear}
Hoyer, P.; Janzing, D.; Mooij, J.~M.; Peters, J.; and Sch{\"o}lkopf, B. 2008.
\newblock Nonlinear causal discovery with additive noise models.
\newblock \emph{Advances in Neural Information Processing Systems}, 21.

\bibitem[{Huang et~al.(2018)Huang, Zhang, Lin, Sch{\"o}lkopf, and
  Glymour}]{huang2018generalized}
Huang, B.; Zhang, K.; Lin, Y.; Sch{\"o}lkopf, B.; and Glymour, C. 2018.
\newblock Generalized score functions for causal discovery.
\newblock In \emph{Proceedings of the 24th ACM SIGKDD International Conference
  on Knowledge Discovery \& Data Mining}, 1551--1560.

\bibitem[{Kumor, Zhang, and Bareinboim(2021)}]{kumor2021sequential}
Kumor, D.; Zhang, J.; and Bareinboim, E. 2021.
\newblock Sequential causal imitation learning with unobserved confounders.
\newblock \emph{Advances in Neural Information Processing Systems}, 34:
  14669--14680.

\bibitem[{Lauritzen(1996)}]{lauritzen1996graphical}
Lauritzen, S.~L. 1996.
\newblock \emph{Graphical models}, volume~17.
\newblock Clarendon Press.

\bibitem[{Loh and Wainwright(2013)}]{loh2013structure}
Loh, P.-L.; and Wainwright, M.~J. 2013.
\newblock Structure estimation for discrete graphical models: Generalized
  covariance matrices and their inverses.
\newblock \emph{The Annals of Statistics}, 3022--3049.

\bibitem[{Lu, Hern{\'a}ndez-Lobato, and Sch{\"o}lkopf(2022)}]{lu2022invariant}
Lu, C.; Hern{\'a}ndez-Lobato, J.~M.; and Sch{\"o}lkopf, B. 2022.
\newblock Invariant causal representation learning for generalization in
  imitation and reinforcement learning.
\newblock In \emph{ICLR2022 Workshop on the Elements of Reasoning: Objects,
  Structure and Causality}.

\bibitem[{Pearl(2009)}]{pearl2009causality}
Pearl, J. 2009.
\newblock \emph{Causality}.
\newblock Cambridge university press.

\bibitem[{Peters, B{\"u}hlmann, and Meinshausen(2016)}]{peters2016causal}
Peters, J.; B{\"u}hlmann, P.; and Meinshausen, N. 2016.
\newblock Causal inference by using invariant prediction: identification and
  confidence intervals.
\newblock \emph{Journal of the Royal Statistical Society: Series B (Statistical
  Methodology)}, 78(5): 947--1012.

\bibitem[{Peters, Janzing, and Sch{\"o}lkopf(2017)}]{peters2017elements}
Peters, J.; Janzing, D.; and Sch{\"o}lkopf, B. 2017.
\newblock \emph{Elements of causal inference: foundations and learning
  algorithms}.
\newblock The MIT Press.

\bibitem[{Pomerleau(1988)}]{pomerleau1988alvinn}
Pomerleau, D.~A. 1988.
\newblock Alvinn: An autonomous land vehicle in a neural network.
\newblock \emph{Advances in Neural Information Processing Systems}, 1.

\bibitem[{Ross and Bagnell(2010)}]{ross2010efficient}
Ross, S.; and Bagnell, D. 2010.
\newblock Efficient reductions for imitation learning.
\newblock In \emph{Proceedings of the Thirteenth International Conference on
  Artificial Intelligence and Statistics}, 661--668. JMLR Workshop and
  Conference Proceedings.

\bibitem[{Ross, Gordon, and Bagnell(2011)}]{ross2011reduction}
Ross, S.; Gordon, G.; and Bagnell, D. 2011.
\newblock A reduction of imitation learning and structured prediction to
  no-regret online learning.
\newblock In \emph{Proceedings of the Fourteenth International Conference on
  Artificial Intelligence and Statistics}, 627--635. JMLR Workshop and
  Conference Proceedings.

\bibitem[{Russell(1998)}]{russell1998learning}
Russell, S. 1998.
\newblock Learning agents for uncertain environments.
\newblock In \emph{Proceedings of the Eleventh Annual Conference on
  Computational Learning Theory}, 101--103.

\bibitem[{Sch{\"o}lkopf et~al.(2021)Sch{\"o}lkopf, Locatello, Bauer, Ke,
  Kalchbrenner, Goyal, and Bengio}]{scholkopf2021toward}
Sch{\"o}lkopf, B.; Locatello, F.; Bauer, S.; Ke, N.~R.; Kalchbrenner, N.;
  Goyal, A.; and Bengio, Y. 2021.
\newblock Toward causal representation learning.
\newblock \emph{Proceedings of the IEEE}, 109(5): 612--634.

\bibitem[{Shimizu et~al.(2006)Shimizu, Hoyer, Hyv{\"a}rinen, Kerminen, and
  Jordan}]{shimizu2006linear}
Shimizu, S.; Hoyer, P.~O.; Hyv{\"a}rinen, A.; Kerminen, A.; and Jordan, M.
  2006.
\newblock A linear non-Gaussian acyclic model for causal discovery.
\newblock \emph{Journal of Machine Learning Research}, 7(10).

\bibitem[{Spencer et~al.(2021)Spencer, Choudhury, Venkatraman, Ziebart, and
  Bagnell}]{spencer2021feedback}
Spencer, J.; Choudhury, S.; Venkatraman, A.; Ziebart, B.; and Bagnell, J.~A.
  2021.
\newblock Feedback in imitation learning: The three regimes of covariate shift.
\newblock \emph{arXiv preprint arXiv:2102.02872}.

\bibitem[{Spirtes(1995)}]{spirtes1995directed}
Spirtes, P. 1995.
\newblock Directed Cyclic Graphical Representations of Feedback Models.
\newblock In \emph{Proceedings of the Eleventh Annual Conference on Uncertainty
  in Artificial Intelligence}, 491--498. Morgan Kaufmann.

\bibitem[{Spirtes et~al.(2000)Spirtes, Glymour, Scheines, and
  Heckerman}]{spirtes2000causation}
Spirtes, P.; Glymour, C.~N.; Scheines, R.; and Heckerman, D. 2000.
\newblock \emph{Causation, prediction, and search}.
\newblock MIT press.

\bibitem[{Sun et~al.(2021)Sun, Wu, Zheng, Liu, Chen, Qin, and
  Liu}]{sun2021recovering}
Sun, X.; Wu, B.; Zheng, X.; Liu, C.; Chen, W.; Qin, T.; and Liu, T.-Y. 2021.
\newblock Recovering latent causal factor for generalization to distributional
  shifts.
\newblock \emph{Advances in Neural Information Processing Systems}, 34:
  16846--16859.

\bibitem[{Taeb and Chandrasekaran(2018)}]{taeb2018interpreting}
Taeb, A.; and Chandrasekaran, V. 2018.
\newblock Interpreting latent variables in factor models via convex
  optimization.
\newblock \emph{Mathematical programming}, 167(1): 129--154.

\bibitem[{Uhler et~al.(2013)Uhler, Raskutti, B{\"u}hlmann, and
  Yu}]{uhler2013geometry}
Uhler, C.; Raskutti, G.; B{\"u}hlmann, P.; and Yu, B. 2013.
\newblock Geometry of the faithfulness assumption in causal inference.
\newblock \emph{The Annals of Statistics}, 436--463.

\bibitem[{Vowels, Camgoz, and Bowden(2022)}]{vowels2022d}
Vowels, M.~J.; Camgoz, N.~C.; and Bowden, R. 2022.
\newblock D’ya like DAGs? A survey on structure learning and causal
  discovery.
\newblock \emph{ACM Computing Surveys}, 55(4): 1--36.

\bibitem[{Wainwright, Jordan et~al.(2008)}]{wainwright2008graphical}
Wainwright, M.~J.; Jordan, M.~I.; et~al. 2008.
\newblock Graphical models, exponential families, and variational inference.
\newblock \emph{Foundations and Trends{\textregistered} in Machine Learning},
  1(1--2): 7--36.

\bibitem[{Xie et~al.(2020)Xie, Cai, Huang, Glymour, Hao, and
  Zhang}]{xie2020generalized}
Xie, F.; Cai, R.; Huang, B.; Glymour, C.; Hao, Z.; and Zhang, K. 2020.
\newblock Generalized independent noise condition for estimating latent
  variable causal graphs.
\newblock \emph{Advances in Neural Information Processing Systems}, 33:
  14891--14902.

\bibitem[{Xie et~al.(2019)Xie, Wu, Liu, Zhong, and Lin}]{xie2019differentiable}
Xie, X.; Wu, J.; Liu, G.; Zhong, Z.; and Lin, Z. 2019.
\newblock Differentiable linearized ADMM.
\newblock In \emph{International Conference on Machine Learning}, 6902--6911.
  PMLR.

\bibitem[{Zhang et~al.(2020{\natexlab{a}})Zhang, Lyle, Sodhani, Filos,
  Kwiatkowska, Pineau, Gal, and Precup}]{zhang2020invariant}
Zhang, A.; Lyle, C.; Sodhani, S.; Filos, A.; Kwiatkowska, M.; Pineau, J.; Gal,
  Y.; and Precup, D. 2020{\natexlab{a}}.
\newblock Invariant causal prediction for block mdps.
\newblock In \emph{International Conference on Machine Learning}, 11214--11224.
  PMLR.

\bibitem[{Zhang et~al.(2020{\natexlab{b}})Zhang, McAllister, Calandra, Gal, and
  Levine}]{zhang2020learning}
Zhang, A.; McAllister, R.; Calandra, R.; Gal, Y.; and Levine, S.
  2020{\natexlab{b}}.
\newblock Learning invariant representations for reinforcement learning without
  reconstruction.
\newblock \emph{arXiv preprint arXiv:2006.10742}.

\bibitem[{Zhang and Ghanem(2018)}]{zhang2018ista}
Zhang, J.; and Ghanem, B. 2018.
\newblock ISTA-Net: Interpretable optimization-inspired deep network for image
  compressive sensing.
\newblock In \emph{Proceedings of the IEEE conference on computer vision and
  pattern recognition}, 1828--1837.

\bibitem[{Zhang, Kumor, and Bareinboim(2020)}]{zhang2020causal}
Zhang, J.; Kumor, D.; and Bareinboim, E. 2020.
\newblock Causal imitation learning with unobserved confounders.
\newblock \emph{Advances in Neural Information Processing Systems}, 33:
  12263--12274.

\bibitem[{Zhang et~al.(2011)Zhang, Peters, Janzing, and
  Sch{\"o}lkopf}]{zhang2011kernel}
Zhang, K.; Peters, J.; Janzing, D.; and Sch{\"o}lkopf, B. 2011.
\newblock Kernel-based conditional independence test and application in causal
  discovery.
\newblock \emph{Proceedings of the Twenty-Seventh Conference on Uncertainty in
  Artificial Intelligence}, 804--813.

\bibitem[{Ziebart et~al.(2008)Ziebart, Maas, Bagnell, Dey
  et~al.}]{ziebart2008maximum}
Ziebart, B.~D.; Maas, A.~L.; Bagnell, J.~A.; Dey, A.~K.; et~al. 2008.
\newblock Maximum entropy inverse reinforcement learning.
\newblock In \emph{Proceedings of the Twenty-Third {AAAI} Conference on
  Artificial Intelligence}, volume~8, 1433--1438. Chicago, IL, USA.

\end{thebibliography}

\end{document}